\documentclass{article}
\usepackage{ijcai24}

\usepackage{times}
\usepackage{soul}
\usepackage{url}
\usepackage[hidelinks]{hyperref}
\usepackage[utf8]{inputenc}
\usepackage[small]{caption}
\usepackage{graphicx}
\usepackage{amsmath}
\usepackage{amsthm}
\usepackage{booktabs}
\usepackage{algorithm}
\usepackage{algorithmic}
\usepackage[switch]{lineno}
\usepackage{amsthm,amsmath,amssymb}
\usepackage{multirow}

\usepackage{mathrsfs}

\title{DeepLight: Reconstructing High-Resolution Observations of Nighttime Light With Multi-Modal Remote Sensing Data}

\author{
Lixian Zhang$^{1,5}$
\and
Runmin Dong$^{2,5,*}$\and
Shuai Yuan$^{3}$\and
Jinxiao Zhang$^{2}$ \and
Mengxuan Chen$^{2}$\and
Juepeng Zheng$^{4}$\and
Haohuan Fu$^{2,5,*}$
\\
\affiliations
$^1$National supercomputing center in Shenzhen, $^2$Tsinghua University\\
$^3$The University of Hong Kong, $^4$Sun-Yat San University\\
$^5$Tsinghua University - Xi’an Institute of Surveying and Mapping Joint Research Center\\
}

\begin{document}

\maketitle

\begin{abstract}

Nighttime light (NTL) remote sensing observation serves as a unique proxy for quantitatively assessing progress toward meeting a series of Sustainable Development Goals (SDGs), such as poverty estimation, urban sustainable development, and carbon emission. However, existing NTL observations often suffer from pervasive degradation and inconsistency, limiting their utility for computing the indicators defined by the SDGs. In this study, we propose a novel approach to reconstruct high-resolution NTL images using multi-modal remote sensing data. To support this research endeavor, we introduce DeepLightMD, a comprehensive dataset comprising data from five heterogeneous sensors, offering fine spatial resolution and rich spectral information at a national scale. Additionally, we present DeepLightSR, a calibration-aware method for building bridges between spatially heterogeneous modality data in the multi-modality super-resolution. DeepLightSR integrates calibration-aware alignment, an auxiliary-to-main multi-modality fusion, and an auxiliary-embedded refinement to effectively address spatial heterogeneity, fuse diversely representative features, and enhance performance in $8\times$ super-resolution (SR) tasks. Extensive experiments demonstrate the superiority of DeepLightSR over 8 competing methods, as evidenced by improvements in PSNR (2.01 dB $ \sim $ 13.25 dB) and PIQE (0.49 $ \sim $ 9.32). Our findings underscore the practical significance of our proposed dataset and model in reconstructing high-resolution NTL data, supporting efficiently and quantitatively assessing the SDG progress. The code and data will be available at https://github.com/xian1234/DeepLight.

\end{abstract}

\begin{figure*}[t]
  \centering
  
  \includegraphics[width=\linewidth]{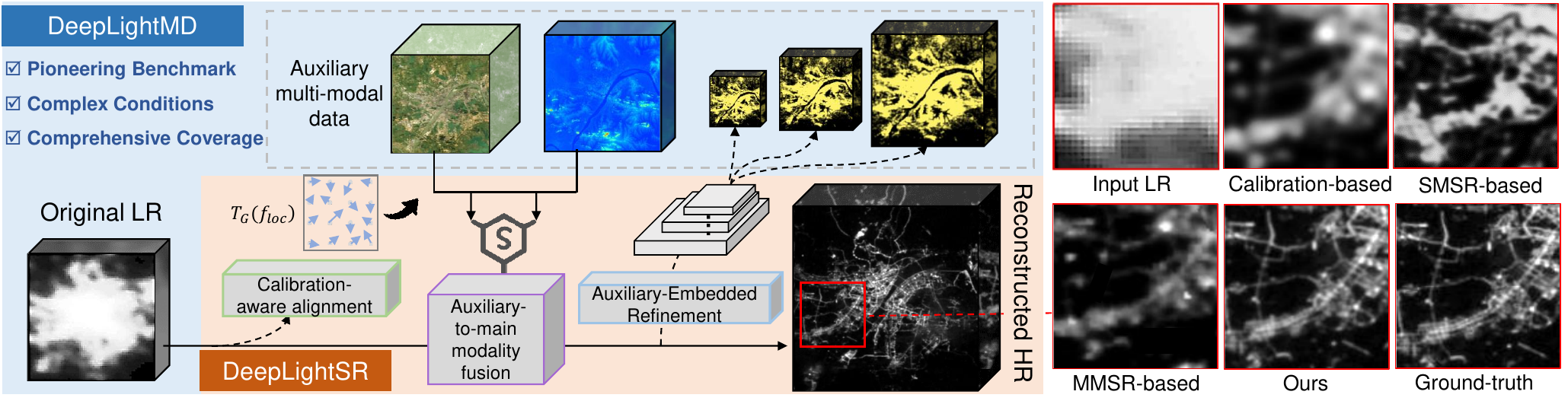}
  \caption{An intuitive example of reconstructing HR NTL image in DeepLightMD using DeepLightSR. The modalities of Daytime Multispectral Observations (DMO), Digital Elevation Model (DEM), and Impervious Surface Product (ISP) are jointly utilized to boost the reconstruction of HR NTL images from LR NTL images.}
  \label{fig1}

\end{figure*}

\section{Introduction} 

The pursuit of the Sustainable Development Goals (SDGs) \cite{sdg} encounters various hurdles such as data insufficiency and lack of quantitative progress assessment \cite{xi2023sd}. Earth observation data from satellites emerges as a promising solution by providing large-scale quantitative attributes \cite{zhang2024swcare} and capturing detailed spatial information essential for understanding Earth's ecosystem, thus being recognized as a highly relevant tool in accomplishing SDGs \cite{guo2023sdgsat}. 

Observations of nighttime light (NTL) over several decades effectively characterize both the extent and intensity of human activities, providing unique insight for quantitatively assessing the progress toward the SDGs \cite{maso2023earth}, such as investigating poverty and inequality \cite{mccallum2022estimating}, monitoring urban sustainable developments \cite{RN35}, and estimating carbon emissions \cite{RN42,RN5}. Due to the lack of onboard calibration, sensor degradation, and satellite orbit drift, the existing historical record of low-resolution NTL suffers from severe quality degradation like blurring, blooming effect, and over-saturation phenomenon \cite{RN94}, highlighting the urgent need for improving the historical NTL data collection. To alleviate these problems, a cost-effective way is to introduce super-resolution (SR) techniques, which reconstruct high-resolution (HR) NTL images from low-resolution (LR) NTL images.

However, as illustrated in Fig. \ref{fig1}, the degradations in historical LR NTL observations are composed of spatially heterogeneous inconsistency, coarse resolution, sensor blurring, atmospheric scattering, etc. The complexity of degradation types makes it a crucial task to improve the quality of LR NTL data for over two decades \cite{RN1000}. Therefore, We present the task of reconstructing HR NTL with multiple modalities in this study. Given an LR NTL image, the algorithm is expected to fuse multi-modality features to guide the reconstruction of calibrated HR NTL image. Nevertheless, the LR images in existing SR datasets are usually manual-simulated with common degradations \cite{dong2023adaptive}. Considering the generalization of the model in addressing various degradations relies on the representativeness of SR datasets, the existing SR datasets are inadequate to support the training for this task. 

To address the aforementioned challenge, we build a pioneering large-scale multi-modality dataset, DeepLightMD, including LR NTL data, HR NTL data, daytime multispectral observations (DMO), digital elevation maps (DEM), and impervious surface products (ISP). DeepLightMD is built to address misalignment, blur, and noise in the $8\times$ reconstruction of HR NTL by incorporating multi-modality data. The involvement of auxiliary modalities has different purposes, with DMO providing finer spatial information and richer spectral information, DEM for spatial consistency, and ISP for importance guidance. A composition of pre-processing methods is introduced to remove noises and geometrical errors in modalities, achieving a geometrically aligned multi-modality dataset. The spatially heterogeneous misalignment and complex inconsistency make DeepLightMD a challenging benchmark dataset for multi-modality super-resolution under complicated degradations. 

To cope with the reconstruction of HR NTL with multi-modality, we propose DeepLightSR, including three modules ($i.e.$, calibration-aware alignment, auxiliary-to-main multi-modality fusion, and auxiliary-embedded refinement) to calibrate modalities under complex inconsistent degradation, address feature fusion of different representative modes, and conduct large-ratio super-resolution tasks. First, the calibration-aware alignment calibrates the main modality and implicitly aligns auxiliary modalities under different representative modes. Second, an auxiliary-to-main multi-modality fusion module extract and distinguish useful low-level features from auxiliary modalities, and then fuse the representative multi-modality features in a stepwise manner to achieve harmonious and aligned features. Third, auxiliary-embedded refinement is proposed to cooperate with auxiliary supervision for detail reconstruction improvement, and multiscale outputs for multi-scale structural information recovering in $8\times$ super-resolution tasks.

We summarize the contributions of this paper as follows: 1) We propose the DeepLightMD, a pioneering national-scale dataset incorporating multi-modal remote sensing data. The DeepLightMD brings a promising way of reconstructing HR NTL images and also a challenge for developing new SR methods capable of addressing complicated degradations in misaligned multi-modality data. 2) We propose DeepLightSR with a calibration-aware alignment, an auxiliary-to-main multi-modality fusion, and an auxiliary-embedded refinement, to build bridges between spatially heterogeneous modality data in reconstucting $\times8$ SR NTL task. 3) Experimental results demonstrate that our method achieves the best results compared with 8 competing methods, highlighting the effectiveness of generating large-scale high-resolution NTL data with our DeepLight SR, supporting efficiently and quantitatively assessing the SDG progress.

\section{Related Work}

Considering that this research is the very first attempt to reconstruct HR NTL data using LR NTL with complex degradation and inconsistency in $\times8$ SR tasks, we focus on investigating the efforts that have been paid (although with a smaller SR scaling factor) or can potentially be applied to achieve improved NTL datasets based on historical NTL records of Defense Meteorological Satellite Program-Operational Linescan System (DMSP-OLS) from both remote sensing (RS) and computer vision (CV) sides.

\subsection{Calibration-Based Methods in Remote Sensing}

Assuming that there is a place where ground truth NTL intensity seldom changed over the past several decades, the main theory of calibration-based methods is to calibrate any pairs of cross-sensor NTL images from one to another using this invariant target. Lots of efforts have been paid to reconstructing calibrated NTL dataset with a single invariant target \cite{RN113}, several invariant targets \cite{RN112}, and invariant regions \cite{RN106}. Apart from expert knowledge, another way relies on determining invariant pixels using machine learning and deep-learning algorithms. For example, A ridge sampling and regression method is proposed to calibrate the NTL time series over the entire globe \cite{RN97}. An embedded convolutional neural network (CNN) framework is proposed to calibrate NTL images \cite{RN120}. Modified U-Nets were introduced to achieve NTL images from different remote sensing modalities \cite{RN121}. However, the existing calibration-based methods are proposed to calibrate NTL images with $\times2$ scaling factor and find it challenging to address most degradation issues in DMSP-OLS NTL data, resulting in limitations of practical and extended applications.

\subsection{Super-Resolution -Based Methods in Computer Vision}

Quantities of deep learning methods have been proposed in effectively conducting super-resolution (SR) tasks in CV \cite{zhang2021making,dong2024building}, such as SRCNN \cite{RN124}, SRResNet \cite{RN123}, and RCAN \cite{RN150}. However, most existing SR methods are designed based on the assumption that LR input and HR target lie in the same domain, which is not valid in this task. We summarize cross-modality SR methods which are or can potentially be applied in applications where LR and HR belong to different domains. The first category is the single-modality SR methods, which do not involve auxiliary modality in the reconstruction of SR results, such as unpaired LR-HR under different domains \cite{RN138}. Besides, several efforts have been paid to different SR targets, such as magnetic resonance (MR) images \cite{RN132}, depth images \cite{RN134}, and thermal images \cite{RN133}. In order to improve the reconstruction performance, auxiliary modalities are introduced to provide useful information to guide the reconstruction. For example, HR RGB visual modality is introduced in the reconstruction of non-visual data \cite{RN135}. JMDL model is proposed to solve the general SR problem with one auxiliary modality \cite{RN137}. However, the existing multi-modality SR methods are designed to alleviate simple degradation and weak misalignment problem in conducting SR tasks, which are difficult to address the complicated degradation and spatially heterogeneous inconsistency of multi-modality NTL SR tasks. 

\section{DeepLightMD}

\subsection{Data Collection}

\begin{figure*}[h]
  \includegraphics[width=\textwidth]{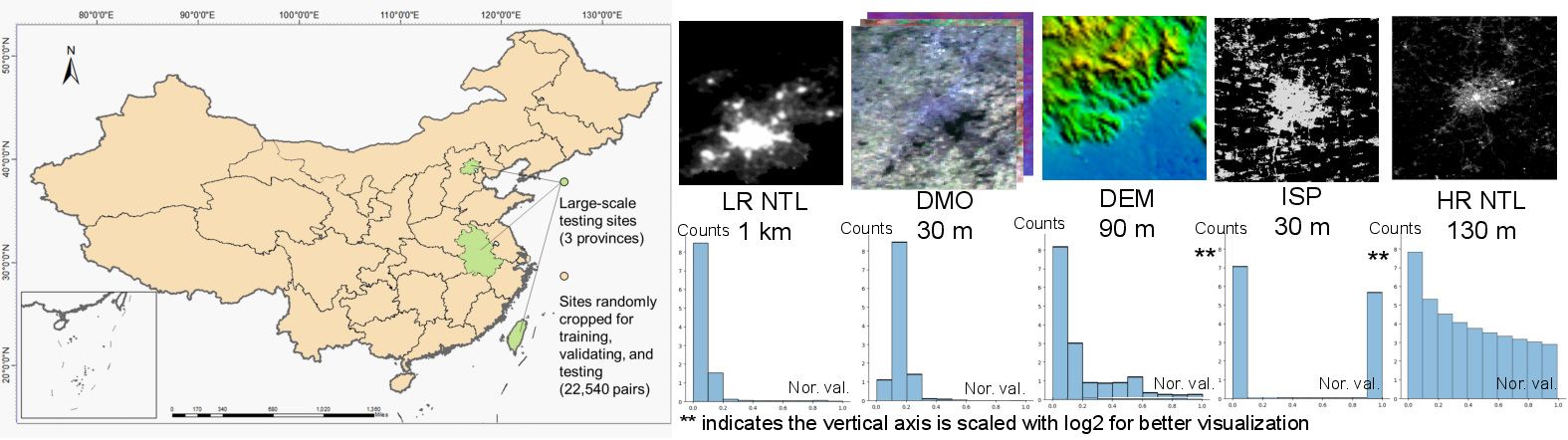}
  \caption{An exploration of the spatial coverage and normalized pixel value distribution histograms of modalities in DeepLightMD.}
  \label{fig2}
\end{figure*}

\noindent\textbf{LR NTL}: Considering that the officially released DMSP-OLS NTL data suffers from interannual inconsistency \cite{RN94}, the Prolonged Artificial Nighttime-light DAtaset (PANDA) \cite{panda} is selected as LR NTL input, which is a long-term DMSP-like NTL dataset of China ranging from 1984-2020. The PANDA can be openly available with longer temporal coverage, better temporal consistency, and improved quality compared to the official DMSP annual composited NTL. The LR NTL data serves as the main modality input, providing information on coarse NTL intensity.

\noindent\textbf{HR NTL}: The composed Luojia NTL data \cite{RN126} is selected as HR NTL in DeepLightMD. The Luojia-01 provides global NTL images at 130 m with 15 days as revisit time. Compared to other commonly used NTL datasets such as VIIRS, the Luojia-01 provides NTL observations with higher spatial resolution, more detailed information, and a wider radiometric range. Besides, the overpass time gap between DMSP and Luojia-01 is much smaller than any other publicly available NTL pairs, reducing the challenges posed by overpass time misalignment \cite{RN7}. 

\noindent\textbf{DMO}: Landsat-8 Operational Land Imager (OLI) is selected as DMO, providing relatively high-resolution (30 m) multispectral images since 2013. The OLI collects data with 8 spectral bands and one panchromatic band at $\sim$ 10 am on each pass with higher spatial resolution, more spectral bands, and a wider radiometric range than NTL data, resulting in much richer features at both global and local scales. We retrieve 7 bands (Band 1-7) of DMO with 30 m resolution of Landsat-8 Top of Atmosphere (TOA) Reflectance imageries. 

\noindent\textbf{DEM}: We select 90 m DEM product releases by Shuttle Radar Topography Mission (SRTM) V4.1 \cite{srtm} in DeepLightMD. The SRTM dataset was collected over an 11-day mission in February 2000, aboard the Space Shuttle Endeavor, and is available for \textasciitilde 80\% of the globe (up to $60^{\circ}$ north and south). The DEM records the elevation information, which is rarely changed over time. The temporal stability and spatial consistency of DEM make it beneficial for improving the spatiotemporal consistency in the reconstruction of NTL data.

\noindent\textbf{ISP}: We select GAIA \cite{RN127} as ISP in this study. The GAIA is retrieved from Landsat series collections using a machine learning algorithm at 30 m resolution. The impervious surface usually indicates the region with constructions, making it an important indicator of human activities. Considering the stable NTL is usually relevant to the impervious surface area, the ISP is selected as auxiliary supervision information to improve the performance of the human settlement region. The impervious surface product of the year 2018 is selected in this research to meet up with the temporal availability of the main modalities. 

All multi-modality remote sensing datasets are processed with normalization, spatial cropping, and random splitting, generating training, testing, and validating materials. Please refer to the supplementary material for details of the data-specific preprocessing. 

\subsection{Dataset Exploration}

\noindent\textbf{Overview}: As illustrated in Fig. \ref{fig2}, DeepLightMD is a national-scale multi-modality SR dataset, consisting of training, validating, and testing materials. The training material contains 18,032 pairs of clips, and the validating material contains 2,254 pairs of samples. To demonstrate evaluation at both local and regional scales, the testing material contains 2,254 pairs of clips and three province-level samples, covering Anhui, Beijing, and Taiwan. Each pair of clips contains LR NTL images ($256\times256$ in size), HR NTL images ($2,048\times2,048$ in size), DMO ($2,048\times2,048$ in size, 7 bands), DEM ($2,048\times2,048$ in size), and ISP ($2,048\times2,048$ in size).

\noindent\textbf{Statistics}: Different from existing multi-modality SR datasets, the modalities in DeepLightMD have huge gaps regarding distributions, radiometric ranges, and spatial resolutions, increasing the difficulties and complexities in model convergence. For example, the DMO data has 7 spectral bands, much more than the others (1 band). The DEM data is in long-tail distribution with a maximum of over 8,000 m, but the majority of pixel values are lower than 1,000. The ISP data is a binary product in which the number of pixels with a value of 0 is 112 times that of pixels with a value of 1. The original LR NTL data is with a radiometric range of 6 bits, smaller than DMO (14 bits) and HR NTL data (12 bits). Considering that the resolution of PANDA is 1 km and the multi-modality super-resolution task is conducted in an $\times8$ super-resolution task, we aim at generating SR NTL images with 125 m resolution in this task.

\noindent \textbf{Highlights}: We summarize the highlights of DeepLightMD as follows: 1) Pioneering Benchmark: DeepLightMD introduces a pioneering benchmark dataset for NTL multi-modality SR tasks, facilitating robust reconstruction of high-resolution NTL imagery and advancing research in both RS and CV domains. 2) Complex Conditions: DeepLightMD presents challenges with complex degradation factors, including spatial misalignment and diverse distributions, demanding models to generalize fusion, alignment, and deblurring techniques for effective performance. As a result, DeepLightMD is more difficult than most existing multi-modality SR datasets. 3) Comprehensive Coverage: DeepLightMD encompasses diverse NTL types across China, spanning various landcover, energy consumption, socioeconomic, and cultural attributes. It features five modalities, from daytime to nighttime observations, capturing natural and human-related measures for comprehensive analysis.

\section{DeepLightSR}
\subsection{Overall Architecture}

\begin{figure*}[h]
  \includegraphics[width=\textwidth]{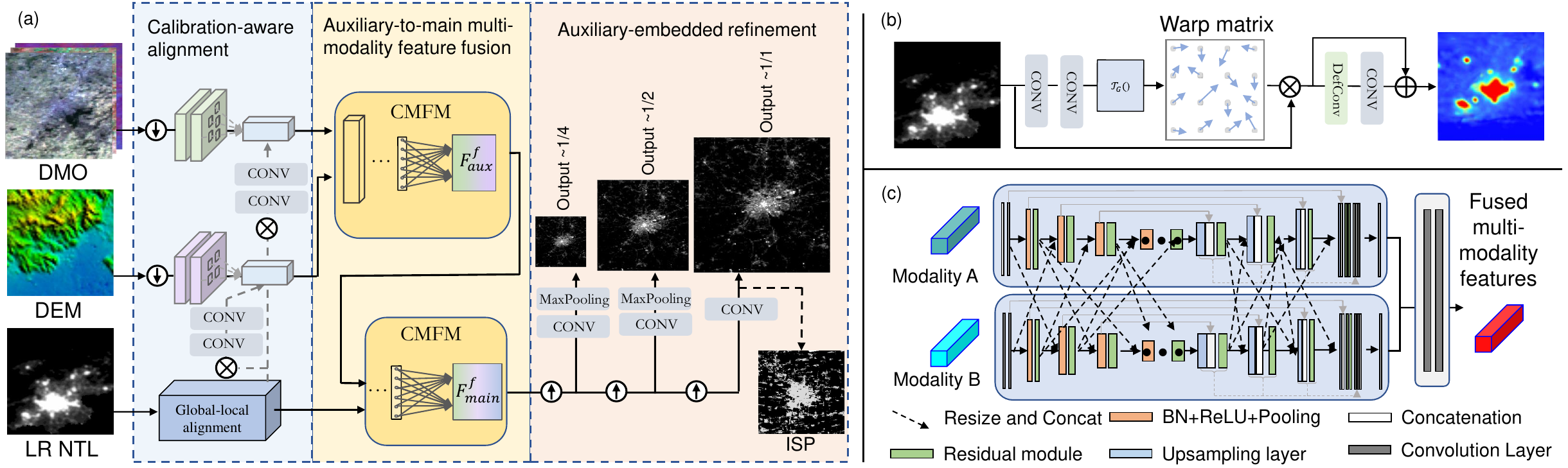}
  \caption{Overview of the proposed DeepLightSR. (a) The overall architecture of DeepLightSR. (b) The proposed global-local alignment in CAA. (c) The proposed cross-modality fusion module in AMFF.}
  \label{fig3}
\end{figure*}

As illustrated in Fig. \ref{fig3}, the DeepLightSR framework partitions multi-modality data into four distinct categories, including main modality input (LR NTL), auxiliary modality inputs (DMO and DEM), main modality supervision (HR NTL), and auxiliary modality supervision (ISP). First, the inputs from both primary and auxiliary modalities undergo calibration-aware alignment to attain aligned features for each modality ($i.e.$, $F_{NTL}$, $F_{DMO}$, and $F_{DEM}$). Second, the aligned features of auxiliary modalities, $F_{DMO}$ and $F_{DEM}$, undergo fusion and combination via a cross-modality fusion module to yield the fused auxiliary feature, $F_{aux}^f$. This fused feature, $F_{aux}^f$, is then integrated with the calibrated $F_{NTL}$ to produce the primary fused feature, $F_{main}^f$. Third, the final SR NTL output is attained under refinement incorporating with multi-scale and auxiliary supervision information in the up-sampling process.

The DeepLightSR is designed in an end-to-end scheme, consisting of two functions $\mathcal{F}_{1} $ and $\mathcal{F}_{2}$.
Given the LR NTL image $ \boldsymbol {N_L} \in \mathbb {R}^{h \times w}$ and the multi-modality input set $ \boldsymbol {M} = { \boldsymbol {M^{DMO}} \in \mathbb {R}^{rh \times rw}, \boldsymbol {M^{DEM}}\in \mathbb {R}^{rh \times rw}}$, $\mathcal{F}{1}$ generates multi-scale SR NTL sets $\boldsymbol {N}{G} ={ \boldsymbol {N}{G,1}, \cdots , \boldsymbol {N}{G,m}}$. Subsequently, $\mathcal{F}{2}$ utilizes the obtained $\boldsymbol {N}{G,m}$ as input to reconstruct the auxiliary SR ISP image $ \boldsymbol {G_{ISP}} \in {0, 1}^{rh \times rw}$.

Overall, DeepLightSR can be optimized with supervision from the HR NTL image $ \boldsymbol{N_H} \in \mathbb {R}^{rh \times rw}$ and auxiliary supervision from the ISP $ \boldsymbol {M_{ISP}} $. Here, $r$, $h$, and $w$ denote the ratio in the SR task, the height, and the width of $ \boldsymbol {N_L} $, respectively. The comprehensive objective function for training DeepLightSR is formulated as follows:

\begin{small}
\begin{equation}
\begin{aligned}
    \boldsymbol {\Theta } = \mathop {\mathrm {argmin}} _{ \boldsymbol {\Theta }} &[ \alpha \mathcal{F}_{1} ||\sum _{j=1}^{m} \beta_{j} (\sum _{i=1}^{n} (\boldsymbol {N}_{L}^{i}, \boldsymbol {M}^{i}; \boldsymbol {\Theta })_{j}-\boldsymbol {N}^{i}_{H,j}||_{1}) + \\
    & (1-\alpha) \mathcal {F}_{2} \sum _{i=1}^{n}(\boldsymbol {M}^{i}_{ISP}\cdot log(\boldsymbol {N}_{G,m}^{i}; \boldsymbol {\Theta }) + \\
    & (1-\boldsymbol {M}^{i}_{ISP})\cdot log(1-(\boldsymbol {N}_{G,m}^{i}; \boldsymbol {\Theta }))],
\end{aligned}
\end{equation}
\end{small}
where $\boldsymbol {\Theta } $ is the parameters of DeepLightSR, the $ n $ and $m$ denote the number of training samples and multiscale outputs, respectively. The $ \boldsymbol {N}_{G,m} \in \mathbb {R}^{rh \times rw} $ represents the obtained SR NTL image in $\mathcal{F}_{1}$. The $\alpha$ and $\beta$ denote the weights of two functions and multiscale tasks.

\subsection{The Calibration-Aware Alignment (CAA)}

In DeepLightMD, the complexity of inconsistency and misalignment across different modality inputs varies significantly, underscoring the necessity of calibrating and aligning each input modality before feature fusion. To this end, the CAA leverages the learned warp parameters to align the severely misaligned primary modality and implicitly guide the alignment of auxiliary modalities. Compared to independently aligning each modality, the proposed calibration-aware strategy optimally utilizes the learned warp for efficient alignment of weakly misaligned modalities and enhances overall model convergence.

The CAA contains three key submodules. First, the auxiliary modalities ($i.e.$, DMO and DEM) $ \boldsymbol {M} $ are rescaled to $ \boldsymbol {M\downarrow} \in \mathbb {R}^{h \times w}$ via learneable downsampling block. The downsampling block comprises several CNN layers with a stride of 2, incorporating learnable parameters to retain essential information while minimizing redundant details during the downscaling process. Second, for the LR NTL input, we propose a global-local alignment (GLA) submodule to generate calibrated NTL feature maps. As illustrated in Fig. \ref{fig3} (b), the GLA contains two components: spatial transformer networks (STN) \cite{RN140} for global alignment and deformable convolutional neural alignment networks (DefCAN) for local alignment. The STN layer employs a localization network $f_{loc}$ and a parameterized grid generator $\mathcal{T}_{G}$ to capture the global warp matrix $\omega$.

\begin{equation}
    \omega = \mathcal{T}_{G}(f_{loc}(\boldsymbol {N}_{L})),
\end{equation}
where the localization network $ f_{loc} $ comprises two CNN layers for regression. The obtained pixel-wise warp matrix $\omega$ is utilized as a global approximation to align the input NTL $\boldsymbol {N}_{L}$. To address the local irregular positions, DefCAN employs a deformable CNN layer (DefConv) \cite{RN139} and a reconstruction CNN layer to generate the aligned NTL features $F_{NTL}$. Third, the DEM and DMO are calibrated with the assistance of the obtained $\boldsymbol {N}_{L}$, leveraging geometric dependencies between modalities and high spatial consistency in DEM data to simplify auxiliary modality alignment. Two additional CNN layers are employed to surpass remaining fractional misalignments between the NTL feature and each auxiliary modality feature. By cooperating with the obtained warp matrix $\omega$, the DMO implicitly "awares" the geometrical deformation information into the calibration process, resulting in coherent and aligned features.

\subsection{Auxiliary-to-main Multi-Modality Feature Fusion (AMFF)}

Significant representativeness gaps exist between different input modality features within DeepLightMD. DMO features ($F_{DMO}$) encapsulate multispectral daytime observations with rich texture details, while DEM features ($F_{DEM}$) offer precise elevation information alongside high-frequency details. On the other hand, LR NTL features ($F_{NTL}$) primarily convey artificial light intensity information in the low-frequency domain. To mitigate cross-modality feature inconsistencies, we adopt a stepwise fusion approach, enabling flexible extraction and differentiation of low-level features while effectively fusing and synthesizing high-level features, thereby enhancing the overall performance in generating fused features ($F_{main}^f$).

Consequently, the fusion of three modalities is simplified into a two-step fusion task. To facilitate efficient fusion of two modality features, we propose the cross-modality fusion module (CMFM) for both auxiliary modality feature fusion and main modality feature fusion tasks. As illustrated in Fig. \ref{fig3} (c), CMFM comprises two parallel branches, each incorporating several repeated residual blocks to progressively extract and fuse features from both modality A and the projected feature from modality B in shallower layers. For deeper layer fusion, the extracted feature from a shallower layer is resized and concatenated into the deeper layer to aid in fusion. This resize-and-concatenate strategy enhances the preservation of global information by providing larger receptive fields, thereby facilitating fusion between modalities with differing representative characteristics and efficient generation of high-level fused features. Subsequently, the extracted modality features are concatenated and further fused using two convolutional blocks to produce the fused multi-modality features.

\subsection{Auxiliary-Embedded Refinement (AER)}
Achieving SR results for large upsampling factors usually suffers from high uncertainty and incompleteness of the missing details and high-frequency information. To boost the reconstruction performance in the $8\times$ SR task, as illustrated in Fig. \ref{fig3} (a), the AER incorporates multi-scale supervision alongside auxiliary supervision to furnish ample guidance for the model in reconstructing intricate textures while mitigating artifact generation during the upsampling process. For multi-scale predictions, we employ three upsampling blocks, each comprising a $1\times1$ point-wise CNN layer followed by a pixel shuffle layer \cite{shi2016real}, facilitating the reconstruction of feature maps into sizes enlarged by a factor of $2\times$. To refine NTL reconstruction across different scales, we compute losses between the output of each upsampling block and the ground truth, which is interpolated to match the same size.

Moreover, as the task entails long-tail distribution regression, reconstructing HR NTL images with L1 loss encounters significant overall underestimation owing to extremely imbalanced distributions (e.g., more than 95\% of pixels exhibit zero NTL intensity in China). Given that NTL applications primarily focus on areas with high human activity and NTL values, an additional CNN layer is introduced to derive ISP $\boldsymbol{G_{ISP}}$ from the obtained $\boldsymbol{N}_{G,m}$. The auxiliary supervision of ISP furnishes crucial guidance to prioritize attention towards manually constructed areas, addressing this imbalance and enhancing model performance.

\begin{figure*}[h]
  \includegraphics[width=\textwidth]{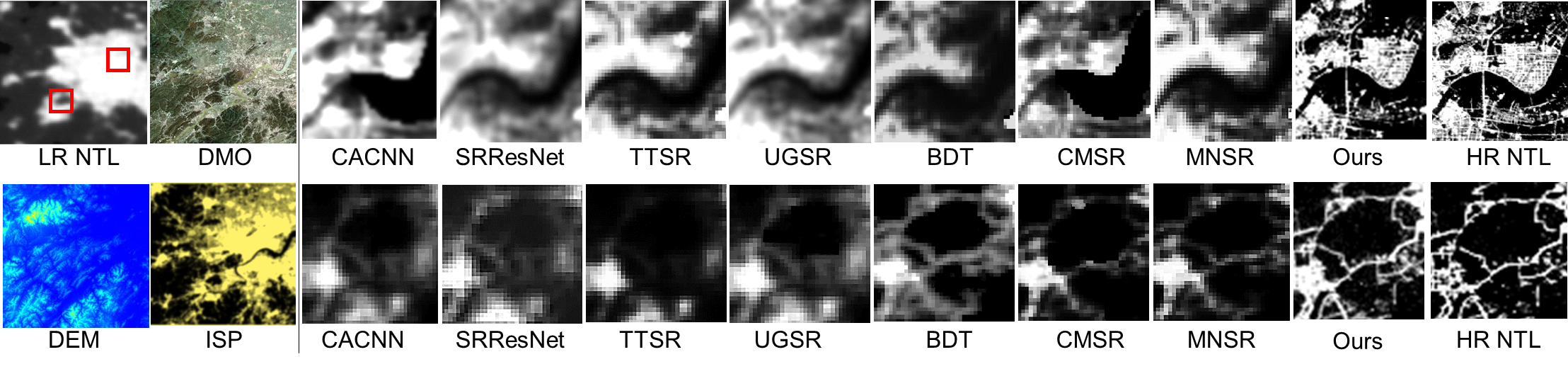}
  \caption{Visual examples for comparison in this study. Five modalities ($i.e.$, LR NTL data, DMO data, DEM data, ISP data, and HR NTL data), eight reconstruction results of competing methods, and the absolute difference between the results of DeepLightSR and HR NTL are presented.}
  \label{fig4}
\end{figure*}

\section{Experiments}
\subsection{Experimental Setups}

\noindent\textbf{Evaluation metrics}: We select 6 metrics for the comprehensive evaluation, including PSNR, SSIM, spectral angle mapper (SAM) \cite{RN143}, universal image quality index (UIQI) \cite{RN142}, correlation coefficient (CC), and perception-based image quality evaluator (PIQE) \cite{RN141}. Apart from the commonly used PSNR, SSIM, and CC, the SAM, UIQI, and PIQE are utilized to evaluate spectral fidelity,  global correlation, and perceptual quality, respectively. 

\noindent \textbf{Competing methods}: We compare the proposed method with 7 representative methods, which can be divided into three categories: calibration-based methods (denoted by “Calibration”): calibration-based auto-encoder CNN (CACNN) \cite{RN120}, single modality super-resolution methods (denoted by “SMSR”): SMSR: SRResNet \cite{RN123}, TTSR \cite{yang2020learning}, UPSR \cite{RN138}, and UGSR \cite{RN133}, and multi-modality super-resolution methods(denoted by “MMSR”): BDT\cite{bdt}, CMSR \cite{RN135} and MNSR \cite{RN136}. Note that the SRResNet and TTSR are trained using pairs of simulated LR NTL downscaled from HR NTL.

\subsection{Experimental Results on DeepLightMD}

\begin{table*}[h]

\caption{Quantitatively comparison on DeepLightMD regarding 6 evaluation metrics.}
\scalebox{0.80}{
\begin{tabular}{lllllllll}
\hline
Method            & Category              & Used modality(s)                  & PSNR $\uparrow$ & SSIM $\uparrow$ & SAM $\downarrow$  & UIQI $\uparrow$  & CC $\uparrow$   & PIQE $\downarrow $    \\ \hline
CACNN \cite{RN120}       & Calibration           & LR NTL + HR NTL                   & 19.14          & 0.7542          & 0.1892          & 0.9105          & 0.9544          & 16.8792         \\ \hline
SRResNet \cite{RN123}     & \multirow{4}{*}{SMSR} & HR NTL                            & 19.63          & 0.7617          & 0.2045          & 0.8826          & 0.9036          & 14.9643         \\
TTSR \cite{yang2020learning}        &                       & HR NTL                            & 21.86          & 0.8129          & 0.1957          & 0.9022          & 0.9163          & 13.1045         \\
UPSR \cite{RN138}        &                       & LR NTL + HR NTL                   & 23.67          & 0.8892          & 0.1451          & 0.9189          & 0.9570          & 9.8691          \\
UGSR \cite{RN133}         &                       & LR NTL + HR NTL                   & 25.68          & 0.9038          & 0.1268          & 0.9269          & 0.9634          & 9.3519          \\ \hline
BDT \cite{bdt}        & \multirow{4}{*}{MMSR} & LR NTL + HR NTL + DMO + DEM + ISP             & 29.31          & 0.9202          & 0.1034          & 0.9545          & 0.9793          & 9.1688          \\
CMSR \cite{RN135}        &            & LR NTL + HR NTL + DMO + DEM + ISP             & 28.68          & 0.9144          & 0.0994          & 0.9577          & 0.9711          & 8.3894          \\
MNSR \cite{RN136}        &                       & LR NTL + HR NTL + DMO + DEM + ISP      & 30.38          & 0.9187          & 0.1083          & 0.9639          & 0.9775          & 8.0422          \\
DeepLightSR (ours) &                       & LR NTL + HR NTL + DMO + DEM + ISP & \textbf{32.39} & \textbf{0.9317} & \textbf{0.0849} & \textbf{0.9717} & \textbf{0.9846} & \textbf{7.5547} \\ \hline
\end{tabular}
}
\label{table1}

\vspace{1em}
\end{table*}

As depicted in Tab. \ref{table1}, our proposed approach consistently outperforms competing methods across all six evaluation metrics, showcasing significant enhancements, particularly in SAM and UIQI metrics. These improvements indicate the ability of our method to preserve both spectral and geometric fidelity in reconstructing SR NTL images. Specifically, while the Calibration method ($ i.e., $ CACNN) excels in CC scores but demonstrates lower performance in pixel-wise evaluation metrics, other Single-Modality SR (SMSR) methods exhibit superior pixel-wise correctness. On the other hand, Multi-Modality SR (MMSR) methods generally achieve superior reconstruction performance, especially in scenarios with severe misalignment. However, both Calibration and SMSR methods fall short in correcting misalignment or achieving spatially precise reconstructions, as illustrated in Fig. \ref{fig4}. By leveraging multi-modality data sources, DeepLightSR achieves substantial improvements in pixel-wise performance compared to competing MMSR methods, demonstrating superior overall performance in addressing inconsistencies and enhancing NTL image quality.

\begin{figure}[h]
  \centering
  \includegraphics[width=\linewidth]{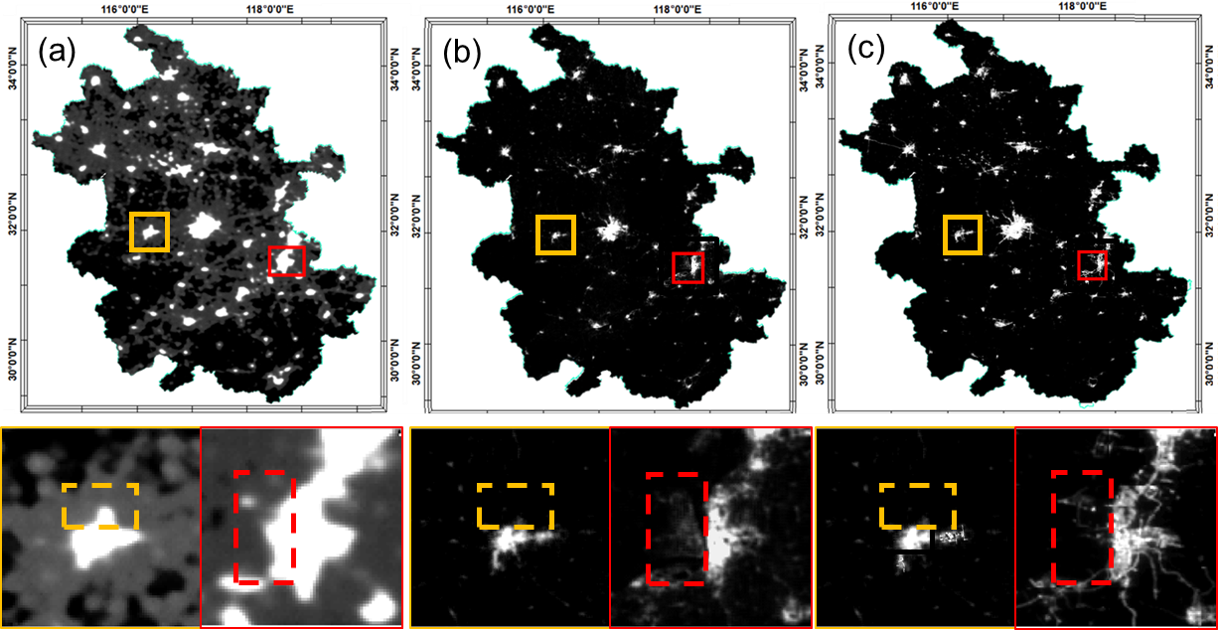}
  \caption{An example of large-scale visual evaluation. (a)-(c) are LR NTL image, SR NTL result of DeepLightSR, and HR NTL image. Dashed boxes are illustrated to better distinguish the effectiveness of DeepLightSR.}
  \label{fig5}
  \vspace{1em}
\end{figure}

To further validate the capabilities of DeepLightSR in large-scale NTL SR tasks, we present regional-scale SR results for three province-level cases. Fig. \ref{fig5} showcases visual evaluations for Anhui province, located in central China. We provide representative examples of zoomed-in areas to illustrate the effectiveness of our method. Notably, the outer shape and intensity of light in areas outside the dashed boxes are aligned with HR NTL images. However, within the dashed boxes, the intensity of light appears excessively high due to over-saturation and blooming effects. Despite these challenges, our super-resolution results demonstrate the method's capability to effectively utilize multi-modality information to correct various reconstruction errors. Absolute errors between retrieved SR NTL images and ground truth further underscore the potential of our method in large-scale multi-modality NTL SR tasks. For additional evaluation results, please refer to the supplementary material.

\subsection{Ablation Study}

\begin{table}[h]
\caption{Experimental results of ablation study.}
\scalebox{0.75}{
\begin{tabular}{lllllll}

\hline
Ablation setups           & PSNR $\uparrow$ & SSIM $\uparrow$ & SAM $\downarrow$  & UIQI $\uparrow$  & CC $\uparrow$   & PIQE $\downarrow$    \\ \hline
w/o LR NTL     & 22.67 & 0.7930  & 0.1739 & 0.9174 & 0.9133 & 12.7153 \\
w/o DMO & 23.17 & 0.8004 & 0.1705 & 0.9025 & 0.9086 & 13.1425 \\
w/o DEM & 27.57 & 0.8455 & 0.1461 & 0.9382 & 0.9529 & 10.6529 \\ \hline    
w/o ISP      & 31.86 & 0.9273 & 0.0975 & 0.9633 & 0.9825 & 9.9297 \\ \hline  
w/o CAA      & 26.56 & 0.8861 & 0.1472 & 0.9586 & 0.9477 & 11.592  \\
w/o AMFF     & 28.62 & 0.8929 & 0.1644 & 0.9475 & 0.9208 & 13.5697 \\
w/o AER & 30.65 & 0.9238 & 0.0919 & 0.9641 & 0.9515 & 10.6388 \\  \hline
Ours full      & \textbf{32.39} & \textbf{0.9317} & \textbf{0.0849} & \textbf{0.9797} & \textbf{0.9846} & \textbf{8.5547}  \\ \hline
\end{tabular}
}
\label{table2}
\vspace{1em}
\end{table} 

\noindent \textbf{Contributions of Multi-Modality Inputs}: We conduct experiments to assess the contribution of each multi-modality input in NTL image reconstruction by training the model with only one or two inputs of multi-modality. We simulate the reduction of specific modalities by replacing the extracted modality features in the cross-modality fusion module with other remaining features. Results, reported in Tab. \ref{table2}, indicate a significant decrease in all evaluation metrics, particularly in CC, when excluding DMO modality, highlighting the valuable insights provided by daytime observations in distinguishing human activity areas from natural ones. Similarly, reducing DEM information as input results in a notable drop in spectral alignment metrics like SAM, emphasizing the crucial role of DEM in maintaining spatial alignment and calibrating other modalities.

\noindent \textbf{Contributions of Auxiliary Supervision Information}: We further investigate the contributions of auxiliary supervision information by conducting experiments without auxiliary supervision ($i.e.$, ISP) during model training. Tab. \ref{table2} reveals performance degradation, especially in perceptual evaluator PIQE, indicating that on-the-fly auxiliary supervision guides the model to focus on reconstruction with improved realistic perception, enhancing its applicability in real-world scenarios. Additionally, since ISP products are retrieved from Landsat series observations, incorporating ISP as a modality input introduces redundancy, leading to decreased overall performance. Hence, ISP serves primarily as auxiliary supervision to enhance the recovery performance of structural information.

\noindent \textbf{Effectiveness of Each Component of the Proposed Method}: Additional experiments are conducted to demonstrate the effectiveness of each component in DeepLightSR. Specifically, replacing CAA removes the cross-modality alignment network, while the setups of other components remain unchanged. Similarly, replacing AMFF removes the connection between input modality networks within the fusion module, and replacing AER eliminates the multiscale output module and corresponding losses. Results in Tab. \ref{table2} show that the calibration-alignment component endows the network with the ability to align spatially irregular inaccuracies for calibrated features. The feature fusion component addresses variance between fused features across network stages, enhancing the model's overall fusion of multi-modality features with different representative modes. Lastly, AER enables the model to discern and align necessary details, improving its capability to address the blooming effect and provide reasonable estimations of NTL intensity.

Please refer to supplementary material for efficiency comparison and more ablation studies, such as multi-scale supervisions and hyper-parameter setups.

\section{Conclusion and Future Work}

In this paper, we introduce DeepLightMD, a pioneering national-scale benchmark dataset encompassing data from five modalities. Building upon this dataset, we propose DeepLightSR, a calibration-aware approach designed to reconstruct HR NTL data. Both qualitative and quantitive results on DeepLightMD demonstrate the effectiveness of the proposed DeepLightSR in generating large-scale, high-resolution NTL data. Furthermore, we identify several avenues for future research with potential societal impact aligned with SDGs: 1) Consistent long-term NTL data reconstruction. There is a growing demand to extend our methodology to reconstruct consistent long-term NTL datasets, addressing temporal variability while ensuring alignment. 2) Incorporating more modalities in the reconstruction of NTL images. Incorporating additional modalities, such as Synthetic Aperture Radar (SAR) and social media data, holds promise in enhancing performance by capturing more nuanced aspects related to terrain and socioeconomic factors.

\section*{Acknowledgements}

This research was supported in part by the National Natural Science Foundation of China (Grant No. T2125006 and No. 42301390), Jiangsu Innovation Capacity Building Program (Project No. BM2022028), China Postdoctoral Science Foundation (Grant No. 2023M731871).

\newpage

\appendix


\end{document}